\documentclass[letterpaper, 10 pt, journal, twoside]{IEEEtran}

\usepackage{style}

\begin{document}
\maketitle
\begin{abstract}
This paper presents Learning-based Autonomous Guidance with RObustness and Stability guarantees (LAG-ROS), which provides machine learning-based nonlinear motion planners with formal robustness and stability guarantees, by designing a differential Lyapunov function using contraction theory. LAG-ROS utilizes a neural network to model a robust tracking controller independently of a target trajectory, for which we show that the Euclidean distance between the target and controlled trajectories is exponentially bounded linearly in the learning error, even under the existence of bounded external disturbances. We also present a convex optimization approach that minimizes the steady-state bound of the tracking error to construct the robust control law for neural network training. In numerical simulations, it is demonstrated that the proposed method indeed possesses superior properties of robustness and nonlinear stability resulting from contraction theory, whilst retaining the computational efficiency of existing learning-based motion planners. 
\end{abstract}
\section{Introduction}
\label{introduction}
\IEEEPARstart{I}{n} the near future of robotic exploration, teams of robots are expected to perform complex decision-making tasks autonomously in extreme environments, where their motions are typically governed by nonlinear dynamics with external disturbances. For such operations to be successful, they need to compute optimal motion plans online while robustly guaranteeing convergence to the target trajectory, both with their limited onboard computational resources. Thus, this work aims to propose a learning-based robust motion planning and control algorithm that meets these challenging requirements.
\subsubsection*{Related Work} 
Learning-based control designs have been an emerging area of research since the rise of neural networks and reinforcement learning~\cite{sutton,ndp}. Model-free approaches learn optimal policies using data obtained in real-world environments, making them robust but not suitable for situations where sampling large training datasets is difficult. Also, proving the robustness and stability properties of such data-driven systems is challenging in general, although some techniques do exist~\cite{boffi2020learning,9029986}. In contrast, model-based methods allow sampling as much data as we want to design the policies by, \eg{}, imitation learning~\cite{glas,9001182}, reinforcement learning~\cite{8593871,NIPS2017_766ebcd5}, or both~\cite{NIPS2016_cc7e2b87,8578338,7995721}. However, the learned controller could yield cascading errors in the real-world environment if its nominal model poorly represents the true underlying dynamics~\cite{cascading}.

Control theoretical approaches to circumvent such difficulties include robust tube-based motion planning~\cite{7989693,ccm,mypaperTAC,ncm,nscm,chuchu,8814758,9303957,sun2021uncertaintyaware,10.1007/BFb0109870,tube_mp,tube_nmp,doi:10.1177/0278364914528132,9290355} equipped with a tracking control law for robustness and stability certificates. Among these are contraction theory-based robust control~\cite{ccm,7989693,mypaperTAC,ncm,nscm,chuchu,8814758,9303957,sun2021uncertaintyaware}, which tracks a target trajectory computed externally by existing motion planners, and thereby robustly keeps the system trajectories in a control invariant tube that satisfies given state constraints. Although these provable guarantees are promising, they still assume that the target trajectory can be computed online solving some planning problems, unlike the aforementioned learning frameworks.
\subsubsection*{Contributions}
\begin{figure*}
    \centering
    \includegraphics[width=170mm]{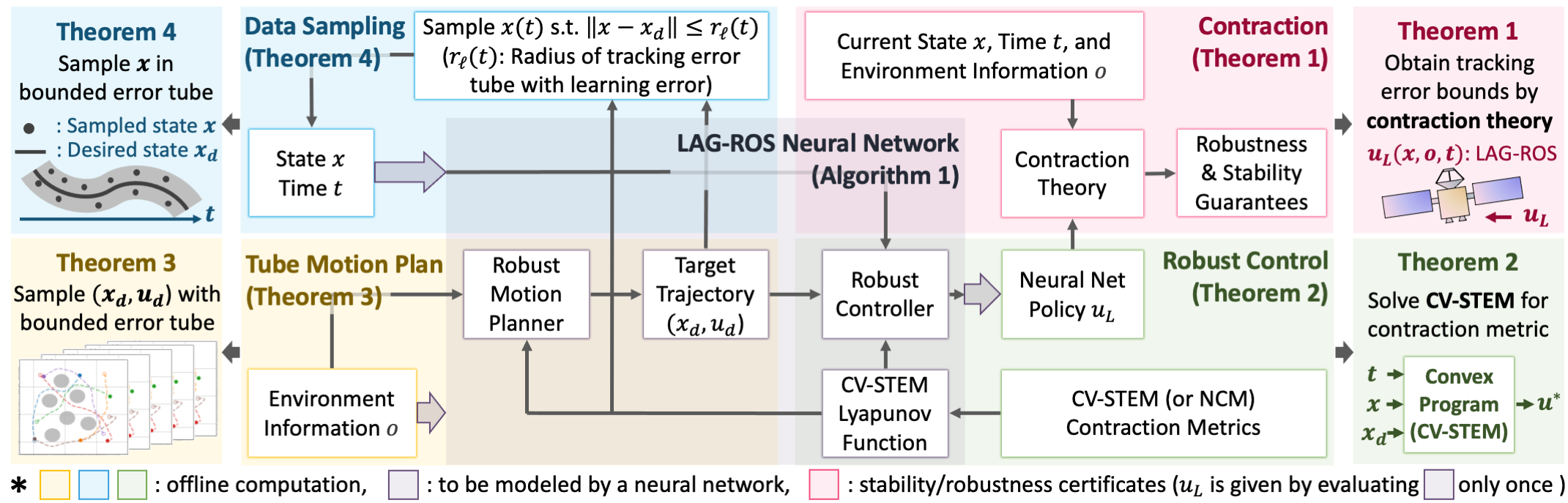}
    \caption{Illustration of LAG-ROS: See Theorems~\ref{Thm:lagros_stability} -- \ref{Prop:sampling} and Algorithm~\ref{lagros_alg} for detailed explanation on each block of the LAG-ROS design. Note that LAG-ROS requires only one neural net evaluation to get $u_L$, and Theorem~\ref{Thm:lagros_stability} provides its robustness and stability guarantees.\label{lagrosdrawing}}
    \vspace{-0.5em}
\end{figure*}
In this paper, we present Learning-based Autonomous Guidance with RObustness and Stability guarantees (LAG-ROS) as a novel way to bridge the gap between the learning-based and robust tube-based motion planners. In particular, while LAG-ROS requires one neural network evaluation to get its control input as in the learning schemes~\cite{8593871,NIPS2017_766ebcd5,9001182,glas,NIPS2016_cc7e2b87,8578338,7995721}, its contraction theory-based architecture still allows obtaining formal robustness and stability guarantees as in~\cite{ccm,7989693,mypaperTAC,ncm,nscm,chuchu,8814758,9303957,sun2021uncertaintyaware}. This framework depicted in Fig.~\ref{lagrosdrawing} is summarized as follows.
\begin{table*}
\caption{Comparison of the Proposed Method with the Learning-based and Robust Tube-based Motion Planners.} \label{learning_summary}
\vspace{-1.5em}
\begin{center}
\begin{tabular}{|l|l|l|l|}
\hline
Motion planning scheme & Policy to be learned & State tracking error $\|x-x_d\|$ & Computational load \\
\hline
\hline 
\ref{itemFF} Learning-based motion planner~\cite{8593871,NIPS2017_766ebcd5,9001182,glas,NIPS2016_cc7e2b87,8578338,7995721} & $(x,o_{\ell},t) \mapsto u_d$ & Increases exponentially (Lemma~\ref{Lemma:naive_learning}) & One neural net evaluation \\
\ref{itemMP} Robust tube-based motion planner~\cite{10.1007/BFb0109870,tube_mp,tube_nmp,doi:10.1177/0278364914528132,9290355,ccm,7989693,mypaperTAC,ncm,nscm,chuchu,8814758,9303957,sun2021uncertaintyaware} & $(x,x_d,u_d,t) \mapsto u^*$ & Exponentially bounded (Theorem~\ref{ncm_clf_thm}) & Computation required to get $(x_d,u_d)$ \\
\ref{itemLAGROS} Proposed method (LAG-ROS) & $(x,o_{\ell},t) \mapsto u^*$ & Exponentially bounded (Theorem~\ref{Thm:lagros_stability}) & One neural net evaluation \\
\hline
\end{tabular}
\end{center}
\vspace{-2.0em}
\end{table*}
\subsubsection{Robustness and Stability Guarantees}
The theoretical foundation of LAG-ROS rests on contraction theory, which utilizes a contraction metric to characterize a necessary and sufficient condition of exponential incremental stability of nonlinear system trajectories~\cite{contraction}. The central result of this paper is that, if there exists a control law which renders a nonlinear system contracting, or equivalently, the closed-loop system has a contraction metric, then LAG-ROS trained to model the controller ensures the Euclidean distance between the target and controlled trajectories to be bounded exponentially with time, linearly in the learning error and size of perturbation. This property helps quantify how small the learning error should be in practice, giving some guidance in choosing design parameters of neural net training. We further show that such a contracting control law and its corresponding contraction metric can be designed explicitly via convex optimization, using the method of CV-STEM~\cite{mypaperTAC,ncm,nscm} to minimize a steady-state upper bound of the LAG-ROS tracking error. 
\subsubsection{State Constraint Satisfaction}
We further exploit the computed bound on the tracking error in generating expert demonstrations for training, so that the learned policy will not violate given state constraints even with the learning error and external disturbances. In this phase, LAG-ROS learns the contracting control law independently of a target trajectory, making it implementable without solving any motion planning problems online unlike~\cite{10.1007/BFb0109870,tube_mp,tube_nmp,doi:10.1177/0278364914528132,9290355,ccm,7989693,mypaperTAC,ncm,nscm,chuchu,8814758,9303957,sun2021uncertaintyaware}. The performance of LAG-ROS is evaluated in cart-pole balancing~\cite{6313077} and nonlinear motion planning of multiple robotic agents~\cite{SCsimulator} in a cluttered environment, demonstrating that LAG-ROS indeed satisfies the formal exponential bound as in~\cite{ccm,7989693,mypaperTAC,ncm,nscm,chuchu,8814758,9303957,sun2021uncertaintyaware} with its computational load as low as that of existing learning-based motion planners~\cite{8593871,NIPS2017_766ebcd5,9001182,glas,NIPS2016_cc7e2b87,8578338,7995721}. In particular, LAG-ROS requires less than $0.1$s for computation in all of these tasks and achieves higher control performances and task success rates (see Sec.~\ref{Sec:performance}), when compared with the existing motion planners in Table~\ref{learning_summary} which outlines the differences of these schemes from our proposed method.
\subsubsection*{Notation}
For $x \in \mathbb{R}^n$ and $A \in \mathbb{R}^{n \times m}$, we let $\|x\|$, $\delta x$, and $\|A\|$, denote the Euclidean norm, infinitesimal variation of $x$, and induced 2-norm, respectively.
For a square matrix $A$, we use the notation $A \succ 0$, $A \succeq 0$, $A \prec 0$, and $A \preceq 0$ for the positive definite, positive semi-definite, negative definite, negative semi-definite matrices, respectively, and $\sym(A) = (A+A^{\top})/2$. Also, $I \in \mathbb{R}^{n\times n}$ denotes the identity matrix.

\section{Learning-based Robust Motion Planning with Guaranteed Stability (LAG-ROS)}
\label{Sec:lagros}
In this paper, we consider the following nonlinear systems with a controller $u \in \mathbb{R}^{m}$ ($f$ \& $B$ are known but $d$ is unknown):
\begin{align}
\label{dynamics}
\dot{x}(t) &= f(x(t),t)+B(x(t),t)u+d(x(t),t)\\
\label{dynamicsd}
\dot{x}_d(o_g,t) &= f(x_d(o_g,t),t)+B(x_d(o_g,t),t)u_d(x_d(o_g,t),o_g,t)
\end{align}
where $t\in \mathbb{R}_{\geq0}$, $f:\mathbb{R}^n\times\mathbb{R}_{\geq0} \mapsto \mathbb{R}^{n}$, $B:\mathbb{R}^n\times\mathbb{R}_{\geq0}\mapsto\mathbb{R}^{n\times m}$, $x:\mathbb{R}_{\geq0} \mapsto \mathbb{R}^{n}$ is the state trajectory of the true system \eqref{dynamics} perturbed by the bounded disturbance $d:\mathbb{R}^{n}\times\mathbb{R}_{\geq0} \mapsto \mathbb{R}^{n}$ \st{} $\sup_{x,t}\|d(x,t)\|=\bar{d}$, $o_g \in \mathbb{R}^g$ is a vector containing global environment information such as initial and terminal states, states of obstacles and other agents, etc., and $x_d:\mathbb{R}^g\times\mathbb{R}_{\geq0}\mapsto\mathbb{R}^{n}$ and $u_d:\mathbb{R}^n\times\mathbb{R}^g\times\mathbb{R}_{\geq0}\mapsto\mathbb{R}^{n}$ are the target trajectories given by existing motion planning algorithms, \eg{}, \cite{10.1007/BFb0109870,tube_mp,tube_nmp,doi:10.1177/0278364914528132,9290355,ccm,7989693,mypaperTAC,ncm,nscm,chuchu,8814758,9303957,sun2021uncertaintyaware}.
\subsection{Problem Formulation of LAG-ROS}
We seek to find $u$ that is computable with one neural network evaluation and guarantees exponential boundedness of $\|x-x_d\|$ in \eqref{dynamics} and \eqref{dynamicsd}, robustly against the learning error and external disturbances. The objective is thus not to develop new learning-based planners that compute $(x_d,u_d)$, but to augment them with formal robustness and stability guarantees. To this end, let us review the following existing planning techniques:
\begin{enumerate}[label={\color{caltechgreen}{(\alph*)}}]
    \item Learning-based motion planner~\cite{glas} or~\cite{8593871,NIPS2017_766ebcd5,9001182,NIPS2016_cc7e2b87,8578338,7995721}:\\
    $(x,o_{\ell}(x,o_g),t) \mapsto u_d(x,o_g,t)$, modeled by a neural network, where $o_{\ell}:\mathbb{R}^n\times\mathbb{R}^g\mapsto\mathbb{R}^{\ell}$ with $\ell \leq g$ is local environment information extracted from $o_g \in \mathbb{R}^g$~\cite{glas}.\label{itemFF}
    \item Robust tube-based motion planner~\cite{7989693} or~\cite{10.1007/BFb0109870,tube_mp,tube_nmp,doi:10.1177/0278364914528132,9290355,ccm,mypaperTAC,ncm,nscm,chuchu,8814758,9303957,sun2021uncertaintyaware}:\\
    $(x,x_d,u_d,t) \mapsto u^*(x,x_d,u_d,t)$, where $u^*$ is a contraction theory-based tracking controller, \eg{}, in Theorem~\ref{ncm_clf_thm}. \label{itemMP}
\end{enumerate}

The robust tube-based motion planner~\ref{itemMP} ensures that the perturbed trajectories $x$ of \eqref{dynamics} stay in an exponentially bounded error tube around the target trajectory $x_d$ of \eqref{dynamicsd}~\cite{ccm,7989693,mypaperTAC,ncm,nscm,chuchu,8814758,9303957,sun2021uncertaintyaware} (see Theorem~\ref{ncm_clf_thm}). However, it requires the online computation of $(x_d,u_d)$ as an input to their control policy, which is not realistic for systems with limited computational resources. 

The learning-based motion planner~\ref{itemFF} circumvents this issue by modeling the target policy $(x,o_{\ell},t) \mapsto u_d$ by a neural network. In essence, our approach, to be proposed in Theorem~\ref{Thm:lagros_stability}, is for providing~\ref{itemFF} with the contraction theory-based stability guarantees~\ref{itemMP}. We remark that~\ref{itemFF} can only assure the tracking error $\|x-x_d\|$ to be bounded by a function which exponentially increases with time, as to be shown in Lemma~\ref{Lemma:naive_learning} for comparison with LAG-ROS of Theorem~\ref{Thm:lagros_stability}.
\subsection{Stability Guarantees of LAG-ROS}
The approach of LAG-ROS bridges the gap between \ref{itemFF} and \ref{itemMP} by ensuring that the distance between the target and controlled trajectories to be exponentially bounded.
\begin{enumerate}[label={\color{caltechgreen}{(\alph*)}},start=3]
    \item Proposed approach (LAG-ROS, see Fig.~\ref{lagrosdrawing}):\\
    $(x,o_{\ell}(x,o_g),t) \mapsto u^*(x,x_d(o_g,t),u_d(x_d(o_g,t),o_g,t),t)$ with $o_{\ell}$ of~\ref{itemFF}, modeled by a neural network $u_L$ of Theorem~\ref{Thm:lagros_stability}, where $u^*$ of \ref{itemMP} is viewed as a function of $(x,o_{\ell},t)$.\label{itemLAGROS}
\end{enumerate}
\begin{theorem}
\label{Thm:lagros_stability}
Suppose that \eqref{dynamics} is controlled to track \eqref{dynamicsd} by the LAG-ROS $u_L=u_L(x,o_{\ell}(x,o_g),t)$, learned to satisfy
\begin{align}
\label{learning_error}
\|u_L-u^*(x,x_d(o_g,t),u_d(x_d(o_g,t),o_g,t),t)\| \leq \epsilon_{\ell},~\forall x,o_g,t    
\end{align}
where $\epsilon_{\ell} \in [0,\infty)$ is the learning error, $u^*$ is the robust control policy of \ref{itemMP} (to be designed in Theorem~\ref{ncm_clf_thm}), and $(x_d,u_d)$ is given by the robust motion planner~\ref{itemMP} (to be discussed in Theorem~\ref{tube_lemma}). Now consider the following virtual system of $y(\mu,t)$ parameterized by $\mu\in[0,1]$, which has $y(\mu=1,t)=x$ of \eqref{dynamics} and $y(\mu=0,t)=x_d$ of \eqref{dynamicsd} as its particular solutions:
\begin{align}
\label{virtual_lagros}
\dot{y} = \zeta(y,x,x_d,u_d,t)+d_y(\mu,x,x_d,u_d,o_g,t)
\end{align}
where $\zeta$ is parameterized by $y$ to verify $\zeta|_{y=x}=f(x,t)+B(x,t)u^*(x,x_d,u_d,t)$ and $\zeta|_{y-x_d}=f(x_d,t)+B(x_d,t)u_d$ (see, \eg{}, \eqref{def:zeta}), $d_y =\mu(B(x,t)(u_L(x,o_{\ell}(x,o_g),t)-u^*(x,x_d,u_d,t))+d(x,t))$, and $x$, $x_d$, $u_d$, and $d$ are as defined in \eqref{dynamics} and \eqref{dynamicsd}. Note that $y(\mu=1,t)=x$ and $y(\mu=0,t)=x_d$ are indeed particular solutions of \eqref{virtual_lagros}. If $\exists \bar{b}\in [0,\infty)$ \st{} $\|B(x,t)\| \leq \bar{b},~\forall x,t$, and if $u^*$ satisfies the following contraction conditions~\cite{Ref:contraction3} with respect to $y$, for a contraction metric $M(y,x,x_d,u_d,t)=\Theta^{\top}\Theta \succ 0$ and $\alpha,\underline{\omega},\overline{\omega} \in (0,\infty)$:
\begin{align}
\label{lagros_contraction}
&\dot{M}+2\sym{}\left(M\frac{\partial \zeta}{\partial y}\right) \preceq -2\alpha M,~\forall y,x,x_d,u_d,t \\
\label{lagros_Mcon}
&\overline{\omega}^{-1}I \preceq M \preceq \underline{\omega}^{-1}I,~\forall y,x,x_d,u_d,t
\end{align}
then we have the following for $\mathtt{e}(t)=x(t)-x_d(o_g,t)$, $\forall o_g$:
\begin{align}
\label{robust_learning_bound}
\|\mathtt{e}(t)\| \leq \mathcal{R}(0)\sqrt{\overline{\omega}}e^{-\alpha t}+\frac{\bar{b}\epsilon_{\ell}+\bar{d}}{\alpha}\sqrt{\frac{\overline{\omega}}{\underline{\omega}}}(1-e^{-\alpha t}) = r_{\ell}(t)
\end{align}
where $\mathcal{R}(t)=\int^{x}_{x_d}\|\Theta\delta y\|$ for $M=\Theta^{\top}\Theta$, $x_d=x_d(o_g,t)$, and $u_d=u_d(x_d(o_g,t),o_g,t)$. 
\end{theorem}
\begin{IEEEproof}
Let $V=\int^{x}_{x_d}\delta y^{\top}M\delta y = \int^{x}_{x_d}\|\Theta\delta y\|^2$ for $(x_d,u_d)$ in \eqref{robust_learning_bound}. Since $\|\partial d_y/\partial \mu\| \leq \bar{b}\epsilon_{\ell}+\bar{d}=\bar{d}_{\epsilon_{\ell}},~\forall \mu,x,o_g,t$, for $d_y$ in \eqref{virtual_lagros} with such $(x_d,u_d)$, the contraction condition \eqref{lagros_contraction} gives
\begin{align}
\dot{V} \leq& \int^{x}_{x_d}\delta y^{\top}\left(\dot{M}+2\sym{}\left(M\frac{\partial \zeta}{\partial y}\right)\right)\delta y+2\bar{d}_{\epsilon_{\ell}}\int^{x}_{x_d}\|M \delta y\| \\
\leq& -2\alpha V+({2\bar{d}_{\epsilon_{\ell}}}/{\sqrt{\underline{\omega}}})\mathcal{R}(t).
\end{align}
Since $d(\|\Theta\delta y\|^2)/dt = 2\|\Theta\delta y\|(d\|\Theta\delta y\|/dt)$, this implies that $\dot{\mathcal{R}} \leq -\alpha\mathcal{R}+{\bar{d}_{\epsilon_{\ell}}}/{\sqrt{\underline{\omega}}}$. Therefore, applying the comparison lemma~\cite[pp.102-103, pp.350-353]{Khalil:1173048} (\ie{}, if $\dot{v}_1 \leq h(v_1,t)$ for $v_1(0) \leq v_2(0)$ and $\dot{v}_2 =h(v_2,t)$, then $v_1(t) \leq v_2(t)$), along with the relation $\mathcal{R}(t) \geq \|\mathtt{e}(t)\|/\sqrt{\overline{\omega}}$, results in \eqref{robust_learning_bound}
\end{IEEEproof}

Theorem~\ref{Thm:lagros_stability} implies that the bound \eqref{robust_learning_bound} decreases linearly in the learning error $\epsilon_{\ell}$ and disturbance $\bar{d}$, and \eqref{dynamics} controlled by LAG-ROS is exponentially stable when $\epsilon_{\ell}=0$ and $\bar{d} = 0$, showing a great improvement over \ref{itemFF} which only gives an exponentially diverging bound as to be derived in Lemma~\ref{Lemma:naive_learning}~\cite{8593871,NIPS2017_766ebcd5,9001182,glas,NIPS2016_cc7e2b87,8578338,7995721}. This property permits quantifying how small $\epsilon_{\ell}=0$ should be to meet the required guidance and control performance, giving some intuition on the neural network architecture (see Sec.~\ref{sec_simulation}). Also, since we model $u^*$ by $u_L(x,o_{\ell},t)$ independently of $x_d$, it is indeed implementable without solving any motion planning problems online unlike the robust motion planners \ref{itemMP}~\cite{10.1007/BFb0109870,tube_mp,tube_nmp,doi:10.1177/0278364914528132,9290355,ccm,7989693,mypaperTAC,ncm,nscm,chuchu,8814758,9303957,sun2021uncertaintyaware}, as outlined in Table~\ref{learning_summary}. If we can sample training data of $u^*$ explicitly considering the bound \eqref{robust_learning_bound}, the LAG-ROS control enables guaranteeing given state constraints even with the learning error $\epsilon_{\ell}$ and external disturbance $d(x,t)$, as will be seen in Sec.~\ref{Sec:contraction}~and~\ref{imitation_learning}.
\begin{remark}
$\epsilon_{\ell}$ of \eqref{learning_error} can only be found empirically in practice, and thus we propose one way to generate training data with \eqref{robust_learning_bound} and use the test error of $u_L$ as $\epsilon_{\ell}$~(see Sec.~\ref{sec_robust_sampling} and Sec.~\ref{sec_simulation}). Note that models $f(x,t)$ learned by system identification for a more accurate description of the nominal dynamics, \eg{}, \cite{8794351}, is still utilizable in \eqref{virtual_lagros} as long as the learned system is contracting~\cite{boffi2020learning}, and the modeling error is bounded. Other types of perturbations, such as stochastic or parametric uncertainty, could be handled using~\cite{nscm,ancm}.
\end{remark}

To appreciate the importance of the guarantees in Theorem~\ref{Thm:lagros_stability}, let us additionally show that~\ref{itemFF}, which models $(x,o_{\ell},t) \mapsto u_d$, only leads to an exponentially diverging bound.
\begin{lemma}
\label{Lemma:naive_learning}
Suppose that $u$ of \eqref{dynamics} is learned to satisfy
\begin{align}
\label{learning_error0}
\|u(x,o_{\ell}(x,o_g),t)-u_d(x,o_g,t)\| \leq \epsilon_{\ell},~\forall x,o_g,t
\end{align}
for $u_d$ of \eqref{dynamicsd} with the learning error $\epsilon_{\ell} \in [0,\infty)$, and that $\exists \bar{b}$ \st{} $\|B(x,t)\| \leq \bar{b},~\forall x,t$. If $f_{c\ell}=f(x,t)+B(x,t)u_d(x,o_g,t)$ is Lipschitz, \ie{}, $\exists L_{f} \in [0,\infty)$ \st{} $\|f_{c\ell}(x_1,o_g,t)-f_{c\ell}(x_2,o_g,t)\| \leq L_f\|x_1-x_2\|,~\forall x_1,x_2 \in \mathbb{R}^n$, then we have the following bound:
\begin{align}
\label{naive_learning_error}
\|\mathtt{e}(t)\| \leq \|\mathtt{e}(0)\|e^{L_{f} t}+L_{f}^{-1}(\bar{b}\epsilon_{\ell}+\bar{d})(e^{L_{f} t}-1).
\end{align}
where $\mathtt{e} = x-x_d$, and $x$, $x_d$, and $\bar{d}$ are given in \eqref{dynamics} and \eqref{dynamicsd}.
\end{lemma}
\begin{IEEEproof}
Integrating \eqref{dynamics} and \eqref{dynamicsd} for $u$ in \eqref{learning_error0} yields $\|\mathtt{e}(t)\| \leq \|\mathtt{e}(0)\|+L_f\int_{0}^t\|\mathtt{e}(\tau)\|d\tau+(\bar{b}\epsilon_{\ell}+\bar{d})t$. Applying the Gronwall-Bellman inequality~\cite[pp. 651]{Khalil:1173048} gives
\begin{align}
    \|\mathtt{e}(t)\| \leq& \|\mathtt{e}(0)\|+\bar{d}_{\epsilon_{\ell}}t+L_{f}\int_{0}^t(\|\mathtt{e}(0)\|+\bar{d}_{\epsilon_{\ell}}\tau)e^{L_{f}(t-\tau)}d\tau
\end{align}
where $\bar{d}_{\epsilon_{\ell}} = \bar{b}\epsilon_{\ell}+\bar{d}$. Thus, integration by parts results in the desired relation \eqref{naive_learning_error}.
\end{IEEEproof}

Lemma~\ref{Lemma:naive_learning} indicates that if there exists either a learning error $\epsilon_{\ell}$ or external disturbance $d$, the tracking error bound grows exponentially with time, and thus \eqref{naive_learning_error} becomes no longer useful for large $t$. Section~\ref{sec_simulation} demonstrates how the computed bounds of \eqref{robust_learning_bound} ($\lim_{t\to \infty}e^{-\alpha t} = 0$) and \eqref{naive_learning_error} ($\lim_{t\to \infty}e^{L_f t} = \infty$) affect the control performance in practice.
\section{Contraction Theory-based Robust and Optimal Tracking Control}
\label{Sec:contraction}
Theorem~\ref{Thm:lagros_stability} is subject to the assumption that we have a contraction theory-based robust tracking control law $u^*$, which satisfies \eqref{lagros_contraction} and \eqref{lagros_Mcon} for a given $(x_d,u_d)$. This section thus delineates one way to extend the method called ConVex Optimization-based Steady-state Tracking Error Minimization (CV-STEM)~\cite{mypaperTAC,ncm,nscm} to find a contraction metric $M$ of Theorem~\ref{Thm:lagros_stability}, which minimizes an upper bound of the steady-state error of \eqref{robust_learning_bound} via convex optimization. Minimizing the bound renders the tube-based planning of Theorem~\ref{tube_lemma} to be given in Sec.~\ref{imitation_learning} for sampling training data less conservative, resulting in a better optimal solution for $(x_d,u_d)$.

In addition, we modify the CV-STEM in~\cite{mypaperTAC} to derive a robust control input $u^*$ which also greedily minimizes the deviation of $u^*$ from the target $u_d$, using the computed contraction metric $M$ to construct a differential Lyapunov function $V = \delta y^{\top}M\delta y$. Note that $u^*$ is to be modeled by a neural network which maps $(x,o_{\ell},t)$ to $u^*$ implicitly accounting for $(x_d,u_d)$ as described in Theorem~\ref{Thm:lagros_stability}, although $u^*$ takes $(x,x_d,u_d,t)$ as its inputs (see Sec.~\ref{sec_robust_sampling}). 
\subsection{Problem Formulation of CV-STEM Tracking Control}
For given $(x_d,u_d)$, we assume that $u^*$ of Theorem~\ref{Thm:lagros_stability} can be decomposed as $u^*(x,x_d,u_d,t) = u_d+K(x,x_d,u_d,t)(x-x_d)$, the generality of which is guaranteed by the following lemma.
\begin{lemma}
\label{u_equivalence_lemma}
Consider a general tracking controller $u$ defined as $u = k(x,x_d,u_d,t)$ with $k(x_d,x_d,u_d,t)=u_d$, where $k:\mathbb{R}^n\times\mathbb{R}^n\times\mathbb{R}^m\times\mathbb{R}_{\geq0}\mapsto\mathbb{R}^{m}$. If $k$ is piecewise continuously differentiable, then $\exists K:\mathbb{R}^n\times\mathbb{R}^n\times\mathbb{R}^m\times\mathbb{R}_{\geq0}\mapsto\mathbb{R}^{m\times n}$ \st{} $u = k(x,x_d,u_d,t) = u_d+K(x,x_d,u_d,t)(x-x_d)$.
\end{lemma}
\begin{IEEEproof}
We have $u=u_d+(k(x,z_d,t)-k(x_d,z_d,t))$ for $z_d=(x_d,u_d)$ due to $k(x_d,z_d,t)=u_d$. Since $k(x,z_d,t)-k(x_d,z_d,t)=\int_0^1(dk(c x+(1-c)x_d,z_d,t)/dc)dc$, choosing $K$ as $\int_0^1(\partial k(c x+(1-c)x_d,z_d,t)/\partial x)dc$ gives the desired relation.
\end{IEEEproof}

Lemma~\ref{u_equivalence_lemma} implies that designing optimal $k$ of $u^* = k(x,x_d,u_d,t)$ reduces to designing optimal $K(x,x_d,u_d,t)$ of $u^*=u_d+K(x,x_d,u_d,t)(x-x_d)$. When \eqref{dynamics} is controlled by the LAG-ROS $u_L$ of Theorem~\ref{Thm:lagros_stability} with such $u^*$, the virtual system of \eqref{dynamics} and \eqref{dynamicsd} which has $y=x$ and $y=x_d$ as its particular solutions can be given by \eqref{virtual_lagros}, where $\zeta$ is defined as follows:
\begin{align}
\label{def:zeta}
\zeta =& \dot{x}_d+(A(x,x_d,u_d,t)+B(x,t)K(x,x_d,u_d,t))(y-x_d)
\end{align}
where $A$ is the State-Dependent Coefficient (SDC) form of the dynamical system \eqref{dynamics} given by Lemma~2 of~\cite{nscm}, which verifies $A(x,x_d,u_d,t)(x-x_d) = f(x,t)+B(x,t)u_d-f(x_d,t)-B(x_d,t)u_d$. Note that $\zeta$ indeed satisfies $\zeta|_{y=x}=f(x,t)+B(x,t)u^*$ and $\zeta|_{y=x_d}=f(x_d,t)+B(x_d,t)u_d$ for such $A$, to have $y=x$ and $y=x_d$ as the particular solutions to \eqref{virtual_lagros}.
\subsection{CV-STEM Contraction Metrics as Lyapunov Functions}
The remaining task is to construct $M$ so that it satisfies \eqref{lagros_contraction} and \eqref{lagros_Mcon}. The CV-STEM approach suggests that we can find such $M$ via convex optimization to minimize an upper bound of \eqref{robust_learning_bound} as $t\to \infty$ when $\alpha$ of \eqref{lagros_contraction} is fixed. Theorem~\ref{ncm_clf_thm} proposes using the metric $M$ designed by the CV-STEM for a Lyapunov function, thereby augmenting $u^*$ with additional optimality to greedily minimize $\|u^*-u_d\|^2$ for $u_d$ in \eqref{dynamicsd}.
\begin{theorem}
\label{ncm_clf_thm}
Suppose that $f$ and $B$ are piecewise continuously differentiable, and let $B=B(x,t)$ and $A=A(x,x_d,u_d,t)$ in \eqref{def:zeta} for notational simplicity. Consider a contraction metric $M(x,x_d,u_d,t) = W(x,x_d,u_d,t)^{-1} \succ 0$ given by the following convex optimization (CV-STEM)~\cite{ncm,mypaperTAC,nscm} to minimize an upper bound on the steady-state tracking error of \eqref{robust_learning_bound}:
\begin{align}
    \label{convex_opt_ncm}
    &{J}_{CV}^* = \min_{\nu>0,\chi \in \mathbb{R},\bar{W}\succ 0} \frac{\bar{b}\epsilon_{\ell}+\bar{d}}{\alpha}\chi \text{~~\st{} \eqref{deterministic_contraction_tilde} and \eqref{W_tilde}}
\end{align}
with the convex constraints \eqref{deterministic_contraction_tilde} and \eqref{W_tilde} given as
\begin{align}
\label{deterministic_contraction_tilde}
&-\dot{\bar{W}}+2\sym{\left(A\bar{W}\right)}-2\nu BR^{-1}B^{\top} \preceq -2\alpha \bar{W},~\forall x,x_d,u_d,t \\
\label{W_tilde}
&I \preceq \bar{W}(x,x_d,u_d,t) \preceq \chi I,~\forall x,x_d,u_d,t
\end{align}
where $\alpha,\underline{\omega},\overline{\omega} \in (0,\infty)$, $\nu = 1/\underline{\omega}$, $\chi = \overline{\omega}/\underline{\omega}$, $\bar{W} = \nu W$, and $R=R(x,x_d,u_d,t)\succ0$ is a weight matrix on the control input. Suppose also that $u^*$ of Theorem~\ref{Thm:lagros_stability} is designed as $u^* = u_d+K^*(x,x_d,u_d,t)\mathtt{e}$, where $\mathtt{e}=x-x_d$, and $K^*$ is found by the following convex optimization for given $(x,x_d,u_d,t)$:
\begin{align}
\label{ncm_robust_control}
&K^* = \text{arg}\min_{K\in \mathbb{R}^{m\times n}}\|u-u_d\|^2 = \text{arg}\min_{K\in \mathbb{R}^{m\times n}}\|K(x,x_d,u_d,t)\mathtt{e}\|^2 \\
\label{stability_clf}
&\text{\st{} } \dot{M}+2\sym{}(MA+MBK(x,x_d,u_d,t)) \preceq -2\alpha M.
\end{align}
Then $M$ satisfies \eqref{lagros_contraction} and \eqref{lagros_Mcon} for $\zeta$ defined in \eqref{def:zeta}, and thus \eqref{robust_learning_bound} holds, \ie{}, we have the exponential bound on the tracking error $\|x-x_d\|$ when the dynamics \eqref{dynamics} is controlled by the LAG-ROS control input $u_L$ of Theorem~\ref{Thm:lagros_stability}. Furthermore, the problem \eqref{ncm_robust_control} is always feasible.
\end{theorem}
\begin{IEEEproof}
Since the differential dynamics of \eqref{virtual_lagros} with \eqref{def:zeta} is given as $\delta \dot{y}=(\partial \zeta/\partial y)\delta y = (A-BK)\delta y$, substituting this into \eqref{lagros_contraction} verifies that \eqref{lagros_contraction} and \eqref{stability_clf} are equivalent. For $\bar{K}=-R^{-1}B^{\top}M$, \eqref{stability_clf} can be rewritten as
\begin{equation}
\nu^{-1}M(-\dot{\bar{W}}+2\sym{}(A\bar{W})-\nu BR^{-1}B^{\top})M \preceq -2\alpha \nu^{-1}M\bar{W}M.
\end{equation}
Since this is clearly feasible as long as $M$ satisfies the condition \eqref{deterministic_contraction_tilde}, this implies that the problem \eqref{ncm_robust_control} is always feasible. Also, multiplying \eqref{lagros_Mcon} by $W_s$ \st{} $W = W_s^2$ with $W_s \succ 0$ from both sides gives \eqref{W_tilde}~\cite{mypaperTAC}. These facts indicate that the conditions \eqref{lagros_contraction} and \eqref{lagros_Mcon} are satisfied for $M$ and $u^*$ constructed by \eqref{convex_opt_ncm} and \eqref{ncm_robust_control}, respectively, and thus we have the exponential bound \eqref{robust_learning_bound} as a result of Theorem~\ref{Thm:lagros_stability}. Furthermore, the problem \eqref{convex_opt_ncm} indeed minimizes an upper bound of \eqref{robust_learning_bound} as $t \to \infty$ due to the relation $0\leq\sqrt{\overline{\omega}/\underline{\omega}} = \sqrt{\chi} \leq \chi$. We remark that \eqref{convex_opt_ncm} is convex as the objective is affine in $\chi$ and \eqref{deterministic_contraction_tilde} and \eqref{W_tilde} are linear matrix inequalities in terms of $\nu$, $\chi$, and $\bar{W}$.
\end{IEEEproof}
\begin{remark}
\label{optimal_remark}
\eqref{convex_opt_ncm} and \eqref{ncm_robust_control} are convex in terms of their respective decision variables and thus can be solved computationally efficiently~\cite[pp. 561]{citeulike:163662}. For systems with a known Lyapunov function (\eg{} Lagrangian systems~\cite[pp. 392]{Slotine:1228283}), we could simply use it to get robust tracking control $u^*$ in Theorem~\ref{ncm_clf_thm} without solving \eqref{ncm_robust_control}, although optimality may no longer be guaranteed in this case. 
\end{remark}
\begin{remark}
The contraction metric construction itself can be performed using a neural network~\cite{ncm,nscm,ancm,chuchu}, leading to an analogous incremental stability and robustness results to those of Theorem~\ref{Thm:lagros_stability}~\cite{ancm,cdc_ncm}.
\end{remark}
\section{Generating Expert Demonstrations}
\label{imitation_learning}
This section presents how to sample training data, \ie{}, the target trajectory $(x_d,u_d)$ of \eqref{dynamicsd} and corresponding perturbed state and CV-STEM robust control $(x,u^*)$, so that $x$ of \eqref{dynamics} controlled by the LAG-ROS control $u_L$ will stay in given admissible state space as long as we have $\|u_L-u^*\| \leq \epsilon_{\ell}$ as in \eqref{learning_error} of Theorem~\ref{Thm:lagros_stability}. To be specific, since $u^*$ of Theorem~\ref{ncm_clf_thm} solves \eqref{convex_opt_ncm} to obtain an optimal error tube around $x_d$, \ie{}, \eqref{robust_learning_bound} of Theorem~\ref{Thm:lagros_stability}, we exploit it in the tube-based motion planning~\cite{7989693} for generating training data, which satisfies given state constraints even under the existence of the learning error $\epsilon_{\ell}$ and disturbance $d$ in \eqref{robust_learning_bound} and \eqref{convex_opt_ncm}. Note that the learning error $\epsilon_{\ell}$ and disturbance upper bound $\bar{d}$ of Theorem~\ref{Thm:lagros_stability} are assumed to be selected \textit{a priori}~\cite{8794351}.
\subsection{Tube-based State Constraint Satisfaction in LAG-ROS}
Given the learning error $\epsilon_{\ell}$ and disturbance upper bound $\bar{d}$ of Theorem~\ref{Thm:lagros_stability}, we sample $(x_d,u_d)$ of \eqref{dynamicsd} by solving the following tube-based global motion planning problem, for each element in a training set containing randomized $o_g$ of \eqref{dynamics}:
\begin{align}
\label{tube_motion_plan}
&\min_{\substack{\bar{x}=\bar{x}(o_g,t)\\\bar{u}=\bar{u}(\bar{x},o_g,t)}}~\int_{0}^{T}c_1\|\bar{u}\|^2+c_2P(\bar{x},\bar{u},t)dt \\
&\text{\st{} $\dot{\bar{x}} = f(\bar{x},t)+B(\bar{x},t)\bar{u}$, $\bar{x}\in\bar{\mathcal{X}}(o_g,t)$, and $\bar{u} \in\bar{\mathcal{U}}(o_g,t)$}
\end{align}
where $c_1\geq0$, $c_2\geq0$, $P(\bar{x},\bar{u},t)$ is some performance-based cost function, $T>0$ is a given time horizon, $\bar{\mathcal{X}}$ is robust admissible state space defined as $\bar{\mathcal{X}}(o_g,t)=\{v(t)\in\mathbb{R}^n|\forall \xi(t) \in \{\xi(t)|\|v(t)-\xi(t)\|\leq r_{\ell}(t)\},~\xi(t)\in\mathcal{X}(o_g,t)\}$, $\mathcal{X}(o_g,t)$ is given admissible state space, $r_{\ell}(t)$ is the right-hand side of \eqref{robust_learning_bound} in Theorem~\ref{Thm:lagros_stability}, and $\bar{u} \in\bar{\mathcal{U}}(o_g,t)$ is an input constraint. The following theorem shows that the LAG-ROS control ensures the perturbed state $x$ of \eqref{dynamics} to satisfy state constraints $x \in \mathcal{X}$, due to the contracting property of $u^*$ in Theorem~\ref{ncm_clf_thm}.
\begin{theorem}
\label{tube_lemma}
If the solution of \eqref{tube_motion_plan} yields $(x_d,u_d)$ of \eqref{dynamicsd}, the LAG-ROS control $u_L$ of Theorem~\ref{Thm:lagros_stability}, where $M$ and $u^*$ are designed by Theorem~\ref{ncm_clf_thm}, ensures the perturbed solution $x(t)$ of \eqref{dynamics} to stay in the admissible state space $\mathcal{X}(o_g,t)$, \ie{}, $x(t) \in \mathcal{X}(o_g,t),\forall t$, even with the learning error $\epsilon_{\ell}$ of $u_L$ in \eqref{learning_error}.
\end{theorem}
\begin{IEEEproof}
Since $x_d \in \bar{\mathcal{X}}$, we have $\xi \in \mathcal{X}$ for all $\xi \in S(x_d) = \{\xi|\|x_d-\xi\|\leq r_{\ell}\}$ by the definition of $\bar{\mathcal{X}}$. Also, if we control \eqref{dynamics} by $u_L$, we have $\|x-x_d\| \leq r_{\ell}$, or equivalently, $x\in S(x_d)$. These two statements indicate that $x \in \mathcal{X}$.
\end{IEEEproof}
\begin{remark}
Theorem~\ref{tube_lemma} implies that if $x_d$ is sampled by \eqref{tube_motion_plan}, the perturbed trajectory \eqref{dynamics} controlled by LAG-ROS $u_L$ of Theorem~\ref{Thm:lagros_stability} will not violate the given state constraints as long as LAG-ROS $u_L$ of Theorem~\ref{Thm:lagros_stability} satisfies \eqref{learning_error}. This helps greatly reduce the need for safety control schemes such as~\cite{8405547}.
\end{remark}
\subsection{Learning Contraction Theory-based Robust Control}
\label{sec_robust_sampling}
To meet the learning error requirement \eqref{learning_error} for the sake of robustness and state constraint satisfaction proposed in Theorems~\ref{Thm:lagros_stability} and~\ref{tube_lemma}, we should also sample $x$ to get training data for the CV-STEM robust control inputs $u^*$ of Theorem~\ref{ncm_clf_thm}.
\begin{figure}
    \centering
    \includegraphics[width=80mm]{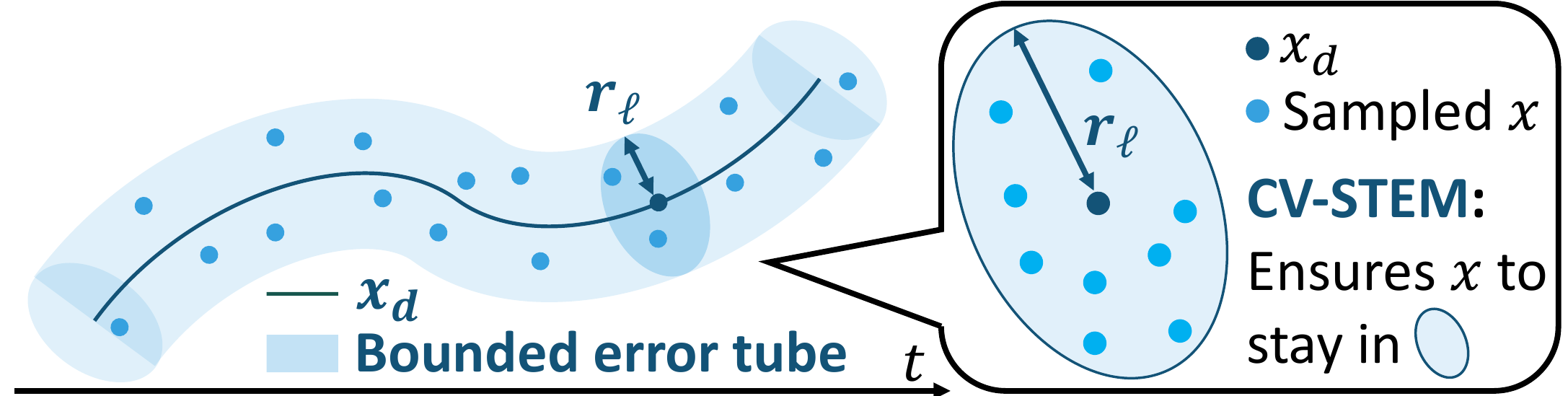}
    \caption{Illustration of state sampling in robust bounded error tube.}
    \label{samplingfig}
    \vspace{-1.0em}
\end{figure}
\begin{theorem}
\label{Prop:sampling}
If \eqref{learning_error} of Theorem~\ref{Thm:lagros_stability} is satisfied $\forall x(t) \in S(x_d) = \{\xi\in\mathbb{R}^n|\|x_d(o_g,t)-\xi\|\leq r_{\ell}(t)\}$ and $\forall o_g,t$ instead of $\forall x,o_g,t$, then $u_L$ with such a constraint still guarantees \eqref{robust_learning_bound}. Also, if we always choose $x(0) = x_d(o_g,0)$ for the perturbed state $x$ in \eqref{dynamics}, the radius of the tube $S(x_d)$ is bounded $\forall o_g,t$.
\end{theorem}
\begin{IEEEproof}
Following the proof of Theorem~\ref{Thm:lagros_stability}, we still get \eqref{learning_error} for all $x(t) \in S(x_d(o_g,t))$, and \eqref{learning_error} implies that the perturbed state $x$ of \eqref{dynamics} indeed lies in $S(x_d(o_g,t))$. When $x(0) = x_d(o_g,0)$, \ie{}, $\mathcal{R}(0)=0$ in \eqref{robust_learning_bound}, $r_{\ell}(t)$ is independent of the state $x$ and thus the radius of the tube $S(x_d)$ is bounded.
\end{IEEEproof}

Using Theorem~\ref{Prop:sampling}, we can sample $x$ in the bounded error tube $S(x_d(o_g,t))$ with $x(0) = x_d(o_g,0)$ to obtain training data for the CV-STEM robust control $u^*$ by \eqref{ncm_robust_control} of Theorem~\ref{ncm_clf_thm}. These samples are to be used for training the LAG-ROS neural network, so we can satisfy the condition \eqref{learning_error} of Theorem~\ref{Thm:lagros_stability} for the pre-defined learning error $\epsilon_{\ell}$.
\begin{remark}
The localization method in~\cite{glas} allows extracting $o_{\ell}$ of~\ref{itemFF} by $o_g$ of \eqref{tube_motion_plan}, to render LAG-ROS applicable to different environments in a distributed way (see Sec.~\ref{sec_simulation}).
\end{remark}

The pseudocode for the offline construction of the LAG-ROS control of Theorem~\ref{Thm:lagros_stability} is presented in Algorithm~\ref{lagros_alg} (see Fig.~\ref{lagrosdrawing} for its visual description). Once we get $u_L(x,o_{\ell},t)$ by Algorithm~\ref{lagros_alg}, $u=u_L$ in \eqref{dynamics} can be easily computed by only evaluating the neural net $u_L$ for a given $(x,o_{\ell},t)$ observed at $(x,t)$, whilst ensuring robustness, stability, and state constraint satisfaction due to Theorems~\ref{Thm:lagros_stability}, \ref{ncm_clf_thm}, and \ref{tube_lemma}.
\begin{algorithm}
\SetKwInOut{Input}{Inputs}\SetKwInOut{Output}{Outputs}
\BlankLine
\Input{Random environment information $\{(o_{g})_i\}^{N}_{i=1}$ \\
Contraction metric $M$ of Theorem~\ref{ncm_clf_thm} \\
Motion planner $\mathcal{P}$ \\ Learning error $\epsilon_{\ell}$ and disturbance bound $\bar{d}$}
\Output{LAG-ROS control $u_L$ of Theorem~\ref{Thm:lagros_stability}}
\For{$i\leftarrow 1$ \KwTo $N$}{
Solve \eqref{tube_motion_plan} of Theorem~\ref{tube_lemma} for $(o_{g})_i$ using $\mathcal{P}$ and obtain a target trajectory and environment observation history $(x_d,u_d,o_g,o_{\ell},t)_i$ \\
Sample $D$ robust CV-STEM control $\{(x,u^*)_{ij}\}_{j=1}^D$ using Theorems~\ref{ncm_clf_thm} and~\ref{Prop:sampling} (see Fig.~\ref{samplingfig})
}
Model $(x,o_{\ell},t)_{ij} \mapsto u^*_{ij}$ by a neural net to satisfy $\|u_L-u^*\| \leq \epsilon_{\ell}$ as in \eqref{learning_error} of Theorem~\ref{Thm:lagros_stability} (see Fig.~\ref{lagrosdrawing})
\caption{LAG-ROS Algorithm}
\label{lagros_alg}
\end{algorithm}
\section{Simulation}
\label{sec_simulation}
Our proposed LAG-ROS framework is demonstrated using multiple motion planning problems under external disturbances ({\color{caltechgreen}\underline{\href{https://github.com/astrohiro/lagros}{https://github.com/astrohiro/lagros}}}). CVXPY~\cite{cvxpy} with the MOSEK solver~\cite{mosek} is used to solve optimization problems in Theorems~\ref{ncm_clf_thm}~and~\ref{tube_lemma} for sampling training data.
\subsection{Simulation Setup}
The maximum admissible control time interval is selected to be $\Delta t_{\rm max} = 0.1$(s). The computational time of each framework is measured for the Macbook Pro laptop (2.2 GHz Intel Core i7, 16 GB 1600 MHz DDR3 RAM), and each simulation result is the average of $50$ simulations for each random environment and disturbance realization.
\subsubsection{Neural Network Training}
\label{Sec:training}
We use a neural network $u_L$ with $3$ layers and $100$ neurons. The network is trained using stochastic gradient descent with training data sampled by Theorems~\ref{ncm_clf_thm}--\ref{Prop:sampling}, and the loss function is defined as $\|u_L-u^*\|$ to satisfy the learning error bound $\epsilon_{\ell}$ of Theorem~\ref{Thm:lagros_stability}. We use $50000$ training samples and select $\epsilon_{\ell} = 0.01$ for all the tasks, but these numbers can be modified accordingly depending on situations, considering the required performance for the error bound \eqref{learning_error} in Theorem~\ref{Thm:lagros_stability}. Note that we can guarantee \eqref{learning_error} only empirically, using the test error computed with a given test set as in standard neural network training.
\subsubsection{Environment Information} $o_g$ of \eqref{dynamicsd} is selected as initial states, target terminal states, and states of obstacles and other agents, if any. In Sec.~\ref{Sec:sc_planning}, the deep set framework~\cite{NIPS2017_f22e4747} is used to extract local information $o_{\ell}$ of \ref{itemFF} from $o_g$ as in~\cite{glas}.
\subsubsection{Performance Measure}
\label{Sec:performance}
The objective function of Theorem~\ref{tube_lemma} for sampling $(x_d,u_d)$ of \eqref{dynamicsd} is selected as $\int_0^T\|\bar{u}\|^2dt$. Since LAG-ROS is not for proposing a new trajectory optimization solver but for augmenting it with the formal guarantees of Theorem~\ref{Thm:lagros_stability}, one could use any other objective, \eg{}, information-based cost~\cite{doi:10.2514/6.2021-1103}, as long as $(x_d,u_d)$ is obtainable.

We define success of each task as the situation where the agent reaches, avoiding collisions, if any, a given target terminal state $x_f$ within a given time horizon $\mathcal{T} \geq T$, \ie{} $\exists t^* \in [0,\mathcal{T}]$ \st{} $\|x(t^*)-x_f\| \leq r_{\ell}(t^*)$ for $r_{\ell}$ in \eqref{robust_learning_bound}, where the value for $r_{\ell}$ is to be defined in the subsequent sections. The success rate is computed as the percentage of successful trials in the total $50$ simulations. Also, we evaluate the performance of each planner, \ref{itemFF}, \ref{itemMP}, and \ref{itemLAGROS} in Sec.~\ref{Sec:lagros}, by the objective function $\int_0^{t^*}\|u\|^2dt$ if the task is successfully completed, and $\int_0^{T}\|u\|^2dt$ otherwise, where $T$ is the nominal time horizon.
\subsubsection{External Disturbances}
As shown in Lemma~\ref{Lemma:naive_learning} and Theorem~\ref{Thm:lagros_stability}, the tracking error bound of learning-based motion planners~\ref{itemFF} increases exponentially with time, whilst it decreases exponentially for the proposed approach~\ref{itemLAGROS}. To make such a difference clear, we consider the situations where $d(x,t)$ of \eqref{dynamics} is non-negligible, and thus~\ref{itemFF} tends to fail the task of Sec~\ref{Sec:performance} due to Lemma~\ref{Lemma:naive_learning}. This is not to argue that \ref{itemFF}~\cite{8593871,NIPS2017_766ebcd5,9001182,glas,NIPS2016_cc7e2b87,8578338,7995721} should be replaced by~\ref{itemLAGROS}, but to imply that they can be improved further to have the guarantees of Theorem~\ref{Thm:lagros_stability}.
\subsubsection{Computational Complexity}
\label{Sec:computation}
Since~\ref{itemFF} and LAG-ROS~\ref{itemLAGROS} are implementable with one neural net evaluation at each $t$, their performance is also compared with~\ref{itemMP} which requires solving motion planning problems to get its control input. Its time horizon is selected to make the trajectory optimization~\cite{47710} solvable online considering the current computational power, for the sake of a fair comparison. We denote the computational time as $\Delta t$ in this section, and it should be less than the maximum control time interval $\Delta t_{\rm max}$, \ie{}, $\Delta t \leq \Delta t_{\rm max} = 0.1$(s).
\subsection{Cart-Pole Balancing}
\label{Sec:cartpole}
We first consider the cart-pole balancing task in Fig.~\ref{lagros_cp_dist_fig} to demonstrate the differences of \ref{itemFF} -- \ref{itemLAGROS} summarized in Table~\ref{learning_summary}. Its dynamics is given in~\cite{ancm,6313077}, and we use $g = 9.8$, $m_c = 1.0$, $m = 0.1$, $\mu_c = 0.5$, $\mu_p = 0.002$, and $l = 0.5$.
\subsubsection{LAG-ROS Training}
$T$, $\mathcal{T}$, and $x_f$ in Sec.~\ref{Sec:performance} are selected as $T=\mathcal{T}=9$(s) and $x_f=[p_f,0,0,0]^{\top}$ for $x$ in Fig.~\ref{lagros_cp_dist_fig}, where $p_f$ is a random terminal position at each episode. We let $\mathcal{R}(0)=0$ and $\bar{d}_{\epsilon} = \bar{b}\epsilon_{\ell}+\bar{d} = 0.75$ in \eqref{robust_learning_bound}, and train the neural net of Sec.~\ref{Sec:training} to have $(\bar{d}_{\epsilon}/{\alpha})\sqrt{\chi} = 3.15$ with $\alpha = 0.60$ using Algorithm~\ref{lagros_alg}. The performance of LAG-ROS \ref{itemLAGROS} is compared with the learning-based planner~\ref{itemFF} and robust tube-based planner~\ref{itemMP}.
\subsubsection{Simulation Results and Discussions}
\begin{figure}
\centerline{
\subfloat{\includegraphics[clip, width=25mm]{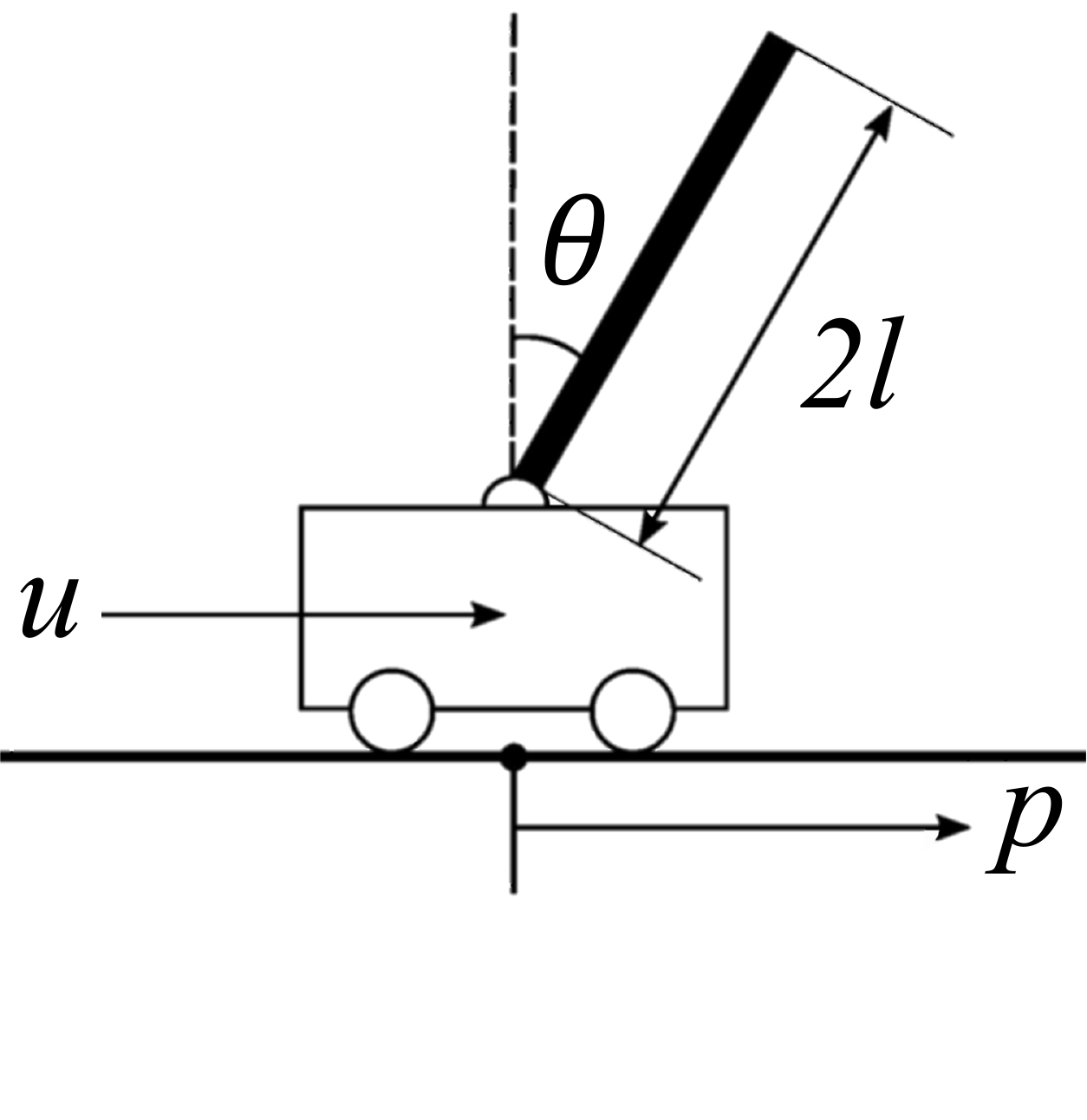}}
\hfil
\subfloat{\includegraphics[clip, width=45mm]{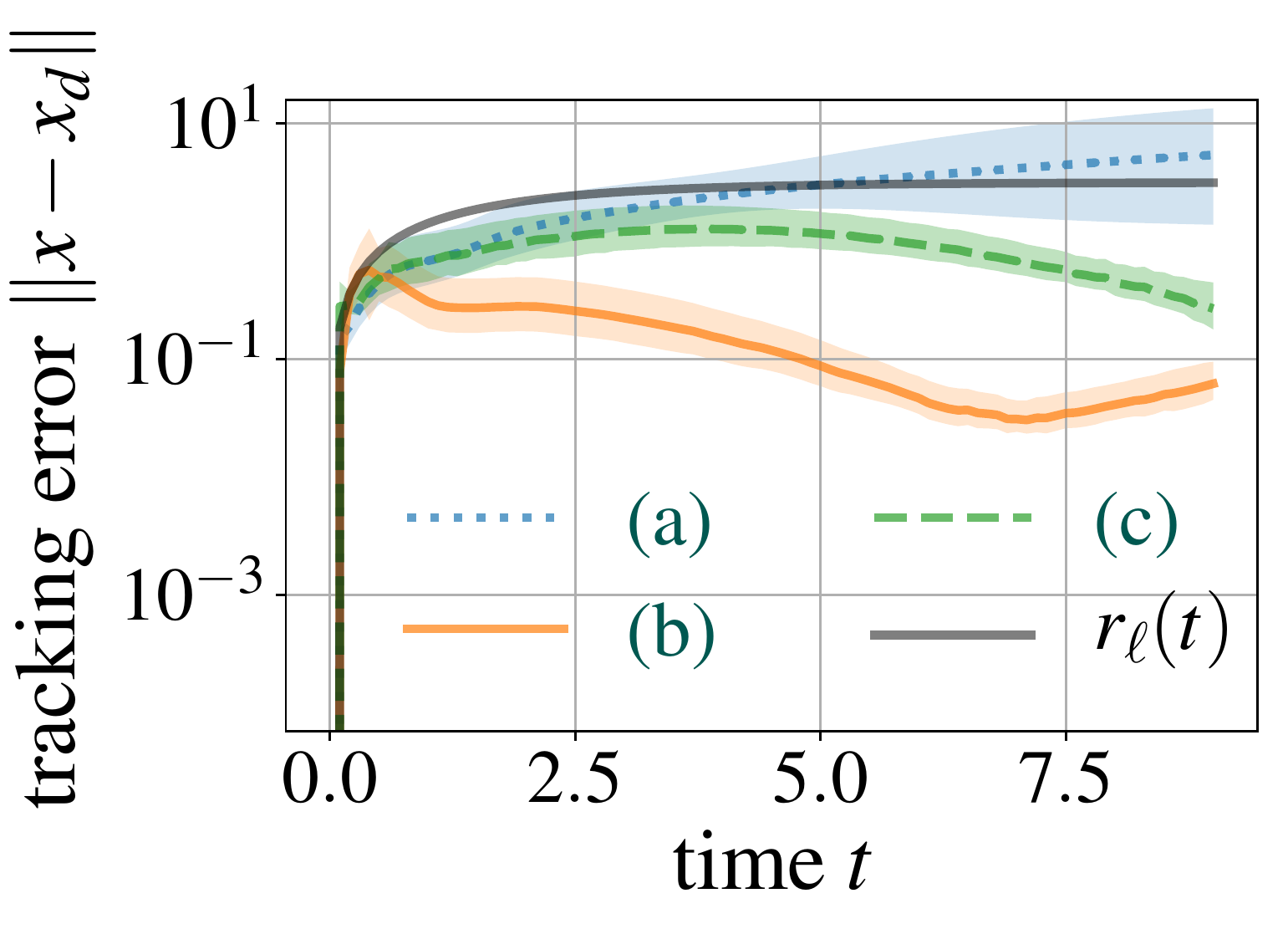}}
}
\vspace{-1.3em}
\caption{Cart-pole balancing task: $x=[p,\theta,\dot{p},\dot{\theta}]^{\top}$, $x$ and $x_d$ are given in \eqref{dynamics} and \eqref{dynamicsd}, \ref{itemFF} -- \ref{itemLAGROS} are given in Sec.~\ref{Sec:lagros}, and $r_{\ell}$ is given in \eqref{robust_learning_bound}. The shaded area denotes the standard deviation ($+1\sigma$ and $-0.5\sigma$).}
\label{lagros_cp_dist_fig}
\vspace{-1.2em}
\end{figure}

Note that the robust tube-based motion planner~\ref{itemMP} (\ie{} $u^*$ of Theorem~\ref{Thm:lagros_stability}) is computable with $\Delta t \leq \Delta t_{\rm max}$ (see Sec.~\ref{Sec:computation}) in this case, and thus we expect that the performance of LAG-ROS~\ref{itemLAGROS} should be worse than that of~\ref{itemMP}, as~\ref{itemLAGROS} is a neural net model that approximates~\ref{itemMP}. The right-hand side of Fig.~\ref{lagros_cp_dist_fig} shows the tracking error $\|x-x_d\|$ of \eqref{dynamics} and \eqref{dynamicsd} averaged over $50$ simulations at each time instant $t$. It is still interesting to see that LAG-ROS of \ref{itemLAGROS} and $u^*$ of \ref{itemMP} indeed satisfies the exponential bound \eqref{robust_learning_bound} of Theorem~\ref{Thm:lagros_stability} given as 
\begin{align}
r_{\ell}(t) = \mathcal{R}(0)\sqrt{\overline{\omega}}+(\bar{d}_{\epsilon}/{\alpha})\sqrt{\chi}(1-e^{-\alpha t}) = 3.15(1-e^{-0.60 t}) \nonumber
\end{align}
for all $t$ with a small standard deviation $\sigma$, unlike learning-based motion planner~\ref{itemFF} with a diverging bound~\eqref{naive_learning_error} and increasing deviation $\sigma$, as can be seen from Fig.~\ref{lagros_cp_dist_fig}. Contraction theory enables such quantitative analysis on robustness and stability of learning-based planners, which is one of the major advantages of our proposed technique.
\renewcommand{\arraystretch}{1.1}
\begin{table}
\caption{Control Performances for Cart-Pole Balancing ($\bar{d}_{\epsilon}=0.75$). \\
Note that the notations are as given in Sec.~\ref{Sec:performance}~and~\ref{Sec:computation}. \label{lagros_cp_tab}}
\vspace{-1.2em}
\begin{center}
\begin{tabular}{|l|c|c|c|c|}
\hline
  & Success rate (\%) & $\int \|u(t)\|^2dt$ & $\Delta t$ (s) \\
\hline
\hline 
\ref{itemFF} & $40.0$ & $4.79 \times 10^2$ & $3.01\times 10^{-2}$ \\
\ref{itemMP} & $100.0$ & $5.19 \times 10^2$ & $1.00\times 10^{-1}$ \\
\ref{itemLAGROS} & $100.0$ & $5.67 \times 10^2$ & $3.01\times 10^{-2}$ \\
\hline
\end{tabular}
\end{center}
\vspace{-2.5em}
\end{table}
\renewcommand{\arraystretch}{1.0}

Table~\ref{lagros_cp_tab} shows the control performance and computational cost of \ref{itemFF} -- \ref{itemLAGROS}, which is a good summary of their differences and trade-offs aforementioned in Sec.~\ref{Sec:lagros} and in Table~\ref{learning_summary}:
\begin{itemize}
    \item \ref{itemFF} approximates $u_d$ of \eqref{dynamics}, and thus requires lower computational cost $\Delta t=0.03$(s) with a smaller objective value $\int \|u\|^2dt=479$, but robustness is not guaranteed (Lemma~\ref{Lemma:naive_learning}) resulting in a $40\%$ success rate.
    \item \ref{itemMP} computes $u^*$ of Theorem~\ref{ncm_clf_thm}, and thus possesses robustness as in Fig.~\ref{lagros_cp_dist_fig} resulting in a $100\%$ success rate, but requires larger $\Delta t=0.1$(s) to compute $x_d$.
    \item \ref{itemLAGROS} approximates $u^*$ independently of $x_d$, and thus possesses robustness of Theorem~\ref{Thm:lagros_stability} as in Fig.~\ref{lagros_cp_dist_fig} resulting in a $100\%$ success rate, even with $\Delta t = 0.03$(s) as small as that of \ref{itemMP}. It yields $9.2\%$ larger $\int \|u\|^2dt$ than \ref{itemMP} as it models $(x,o_{\ell},t) \mapsto u^*$, not $(x,o_{\ell},t)\mapsto u_d$.
\end{itemize}
It is demonstrated that LAG-ROS indeed possesses the robustness and stability guarantees of $u^*$ as in Theorem~\ref{Thm:lagros_stability}, unlike~\ref{itemFF}, with significantly lower computational cost than that of~\ref{itemMP} as expected.
\subsection{Multi-Agent Nonlinear Motion Planning}
\label{Sec:sc_planning}
The advantages of~\ref{itemLAGROS} demonstrated in Sec.~\ref{Sec:cartpole} are more appreciable in the problem settings where the robust tube-based motion planner~\ref{itemMP} is no longer capable of computing a global solution with $\Delta t \leq \Delta t_{\rm max}$ (see Sec.~\ref{Sec:computation}). We thus consider motion planning and collision avoidance of multiple robotic simulators~\cite{SCsimulator} in a cluttered environment with external disturbances, where the agents are supposed to perform tasks based only on local observations $o_{\ell}$ of \ref{itemFF}. Note that its nonlinear equation of motion is given in~\cite{SCsimulator} and all of its parameters are normalized to $1$.  
\subsubsection{LAG-ROS Training}
We select $T$, $\mathcal{T}$, and $x_f$ in Sec.~\ref{Sec:performance} as $T=30$(s), $\mathcal{T}=45$(s), and $x_f=[p_{xf},p_{yf},0,0]^{\top}$, where $(p_{xf},p_{yf})$ is a random position in $(0,0) \leq (p_{xf},p_{yf}) \leq (5,5)$. We let $\mathcal{R}(0)=0$, and train the neural network of Sec.~\ref{Sec:training} by Algorithm~\ref{lagros_alg} to get $(\bar{d}_{\epsilon}/{\alpha})\sqrt{\chi} = 0.125$ with $\alpha = 0.30$ in \eqref{robust_learning_bound}. In particular, the following error tube \eqref{robust_learning_bound}:
\begin{align}
r_{\ell}(t) = (\bar{d}_{\epsilon}/{\alpha})\sqrt{\chi}(1-e^{-\alpha t}) = 0.125(1-e^{-0.30 t}) \nonumber
\end{align}
is used for sampling $(x_d,u_d)$ by Theorem~\ref{tube_lemma} with an input constraint $u_i\geq 0,~\forall i$ to avoid collisions with a random number of multiple circular obstacles ($0.5$m in radius) and of other agents, even under the learning error and disturbances. When training LAG-ROS, the input constraint is satisfied by using a ReLU function for the network output, and the localization technique~\cite{glas} is used to extract $o_{\ell}$ of~\ref{itemFF} from $o_g$ of \eqref{dynamics} for its distributed implementation, with the communication radius $2.0$m. Its performance is compared with~\ref{itemFF},~\ref{itemMP}, and a centralized planner~\ref{itemC} which is not computable with $\Delta t \leq \Delta t_{\rm max}$, where $\Delta t$ is given in Sec.~\ref{Sec:computation}:
\begin{enumerate}[label={\color{caltechgreen}{(\alph*)}},start=4]
    \item Centralized robust motion planner:\\
    $(x,x_d,u_d,t) \mapsto u^*$, offline centralized solution of \ref{itemMP}. \label{itemC}
\end{enumerate}
\subsubsection{Remarks on Sub-optimal Trajectories}
Multi-agent problems have sub-optimal solutions with optimal values close to the global optimum~\cite{8424034}, and thus~\ref{itemLAGROS} does not necessarily take the same $x_d$ as that of the centralized planner~\ref{itemC} in the presence of disturbances, as depicted in Fig.~\ref{lagros_sc_dist_fig}. However, it implicitly guarantees tracking to its own (sub-)optimal $x_d$ due to Theorem~\ref{Thm:lagros_stability}, which means we can still utilize their objective value $\int\|u\|^2dt$ and success rate to evaluate their performance.
\subsubsection{Implication of Simulation Results}
\label{Sec:implication}
Figure~\ref{lagros_sc_fig} shows one example of the trajectories of the motion planners \ref{itemFF} -- \ref{itemC}, under external disturbances with $\bar{d} = \sup_{x,t}\|d(x,t)\| = 0.4$.
\begin{itemize}
    \item For~\ref{itemFF}, the tracking error accumulates exponentially with time due to $\bar{d}$ (Lemma~\ref{Lemma:naive_learning}), which necessitates the use of safety control for avoiding collisions~\cite{glas}. Such non-optimal control inputs also increase the error in \eqref{naive_learning_error} as \ref{itemFF}~does not possess any robustness guarantees.
    \item For~\ref{itemMP}, robustness is guaranteed by Theorem~\ref{ncm_clf_thm} but can only obtain locally optimal $(x_d,u_d)$ of \eqref{dynamicsd}, as its time horizon has to be small enough to make the problem solvable within $\Delta t \leq \Delta t_{\rm max} = 0.1$(s), and the agent only has access to local information. This renders some agents stuck in local minima as depicted in Fig.~\ref{lagros_sc_fig}.
    \item LAG-ROS \ref{itemLAGROS} tackles these two problems by providing formal robustness and stability guarantees of Theorems~\ref{Thm:lagros_stability} -- \ref{Prop:sampling}, whilst implicitly knowing the global solution (only from the local information $o_{\ell}$ as in~\cite{glas}) without computing it online. It satisfies the given state constraints due to Theorem~\ref{tube_lemma} as can be seen from Fig.~\ref{lagros_sc_fig}
\end{itemize}
\subsubsection{Simulation Results and Discussions}
\begin{figure*}
    \centering
    \includegraphics[width=125mm]{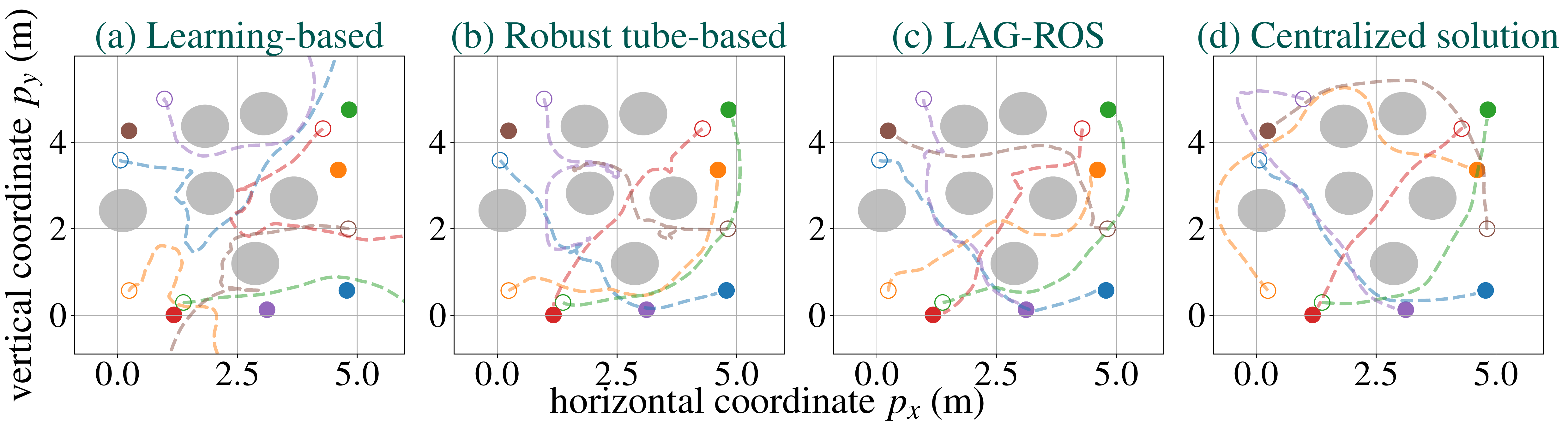}
    \vspace{-1.2em}
    \caption{Trajectories for the learning-based planner~\ref{itemFF}, robust tube-based planner~\ref{itemMP}, LAG-ROS~\ref{itemLAGROS}, and offline centralized solution~\ref{itemC} ($\circ$: start, $\bullet$: goal).}
    \label{lagros_sc_fig}
\vspace{-1.3em}
\end{figure*}

\begin{figure}
    \centering
    \includegraphics[width=80mm]{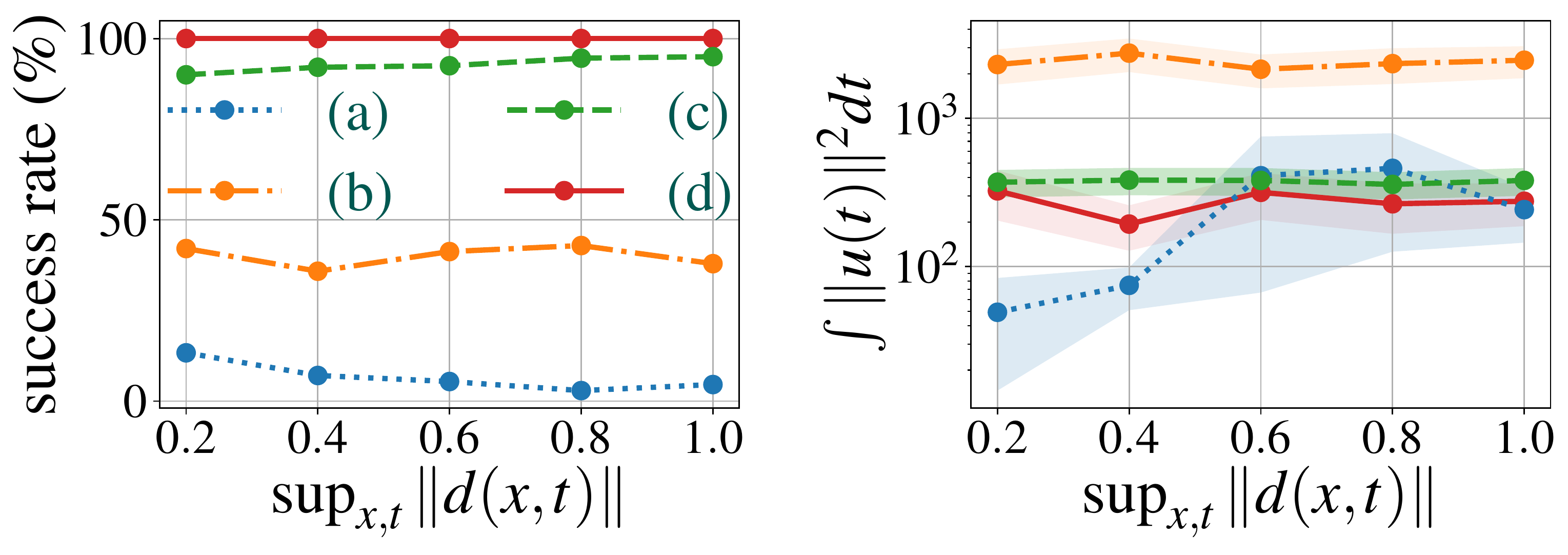}
    \vspace{-1.2em}
    \caption{Control performances versus $\sup_{x,t}\|d(x,t)\|$ of \eqref{dynamics} for multi-agent nonlinear motion planning. Note that \ref{itemFF} -- \ref{itemC} are as given in Fig.~\ref{lagros_sc_fig}, and the performances are as defined in Sec.~\ref{Sec:performance}. The shaded area denotes the standard deviation ($\pm10^{-1}\sigma$).}
    \label{lagros_sc_dist_fig}
\vspace{-1.5em}
\end{figure}

Figure~\ref{lagros_sc_dist_fig} and Table~\ref{lagros_sc_tab} summarize the simulation results which corroborate the arguments of Sec.~\ref{Sec:implication} implied by Fig.~\ref{lagros_sc_fig}:
\begin{itemize}
    \item \ref{itemFF} satisfies the requirement on the computational cost since we have $\Delta t \leq \Delta t_{\rm max} = 0.1$(s) as shown in Table~\ref{lagros_sc_tab}, but its success rate remains the lowest for all $\bar{d}=\sup_{x,t}\|d(x,t)\|$ due to the cumulative error \eqref{naive_learning_error} of Lemma~\ref{Lemma:naive_learning}, as can be seen in Fig.~\ref{lagros_sc_dist_fig} and Table~\ref{lagros_sc_tab}. Although its objective value is the smallest for $\bar{d} \leq 0.6$ since it models $u_d$, it gets larger for larger $\bar{d}$, due to the lack of robustness to keep $x$ around $x_d$.
    \item \ref{itemMP}~has a success rate higher than that of~\ref{itemFF}, but still lower than $50$\% as it can only compute sub-optimal $x_d$ under the limited computational capacity. In fact, we have $\Delta t \geq \Delta t_{\rm max}$ in this case, which means it requires a slightly better onboard computer. Also, it uses excessive control effort larger than $10^3$ due to such sub-optimality.
    \item \ref{itemLAGROS} achieves more than $90$\% success rates for all $\bar{d}$, and its objective value remains only $1.64$ times larger than that of the centralized planner \ref{itemC}, even without computing $(x_d,u_d)$ of \eqref{dynamicsd} online. Its computational cost is as low as that of \ref{itemFF}, satisfying $\Delta t \leq \Delta t_{\rm max} = 0.1$(s) while retaining a standard deviation smaller than~\ref{itemFF}.
\end{itemize}
These results imply that LAG-ROS indeed enhances learning-based motion planners~\ref{itemFF} with robustness and stability guarantees of contraction theory as in \ref{itemC} (and \ref{itemMP}), thereby bridging the technical gap between them.
\renewcommand{\arraystretch}{1.2}
\begin{table}
\caption{Computational time of each motion planner for Multi-agent Nonlinear Motion Planning.\label{lagros_sc_tab}}
\vspace{-1.5em}
\begin{center}
\begin{tabular}{|l|c|c|c|c|}
\hline
  & \ref{itemFF} & \ref{itemMP} & \ref{itemLAGROS} & \ref{itemC} \\
\hline
\hline 
$\Delta t$ (s) & $4.63\times 10^{-2}$ & $1.53\times 10^{-1}$ & $4.66\times 10^{-2}$ & $1.12\times 10^{3}$ \\
\hline
\end{tabular}
\end{center}
\vspace{-2.0em}
\end{table}
\renewcommand{\arraystretch}{1.0}
\section{Conclusion}
\label{sec_conclusion}
In this work, we propose a new learning-based motion planning framework, called LAG-ROS, with the formal robustness and stability guarantees of Theorem~\ref{Thm:lagros_stability}. It extensively utilizes contraction theory to provide an explicit exponential bound on the distance between the target and controlled trajectories, even under the existence of the learning error and external disturbances. Simulation results demonstrate that it indeed satisfies the bound in practice, thereby yielding consistently high success rates and control performances in contrast to the existing motion planners~\ref{itemFF} and~\ref{itemMP}. Note that other types of disturbances can be handled, using~\cite{nscm} for stochastic systems and~\cite{ancm} for parametric uncertain systems.
\subsubsection*{Acknowledgments}
This work was in part funded by the Jet Propulsion Laboratory, California Institute of Technology, and benefited from discussions with J. Castillo-Rogez, M. D. Ingham, and J.-J. E. Slotine.

\bibliographystyle{IEEEtran}
\bibliography{root}

\end{document}